%% file: main.tex
\documentclass[runningheads]{llncs}

\usepackage{accv}

\usepackage{accvabbrv}

\usepackage{graphicx}
\usepackage{booktabs}

\usepackage[accsupp]{axessibility}  

\usepackage{hyperref}

\usepackage{orcidlink}
\usepackage{wrapfig}
\usepackage{graphicx}
\usepackage{booktabs}
\usepackage{multirow}

\begin{document}

\title{Language-Guided Joint Audio-Visual Editing via One-Shot Adaptation} 

\titlerunning{Language-Guided Joint Audio-Visual Editing}

\author{Susan Liang\inst{1}\and
Chao Huang\inst{1}\and
Yapeng Tian\inst{1}\and \\
Anurag Kumar\inst{2} \and
Chenliang Xu\inst{1}
}

\authorrunning{S. Liang et al.}

\institute{University of Rochester, Rochester NY 14627, USA \and
Meta Reality Labs Research, Redmond WA 98052, USA\\}

\maketitle

\input{sec/0_abstract}    
\begin{figure}[t]
    \centering
    \includegraphics[width=\linewidth]{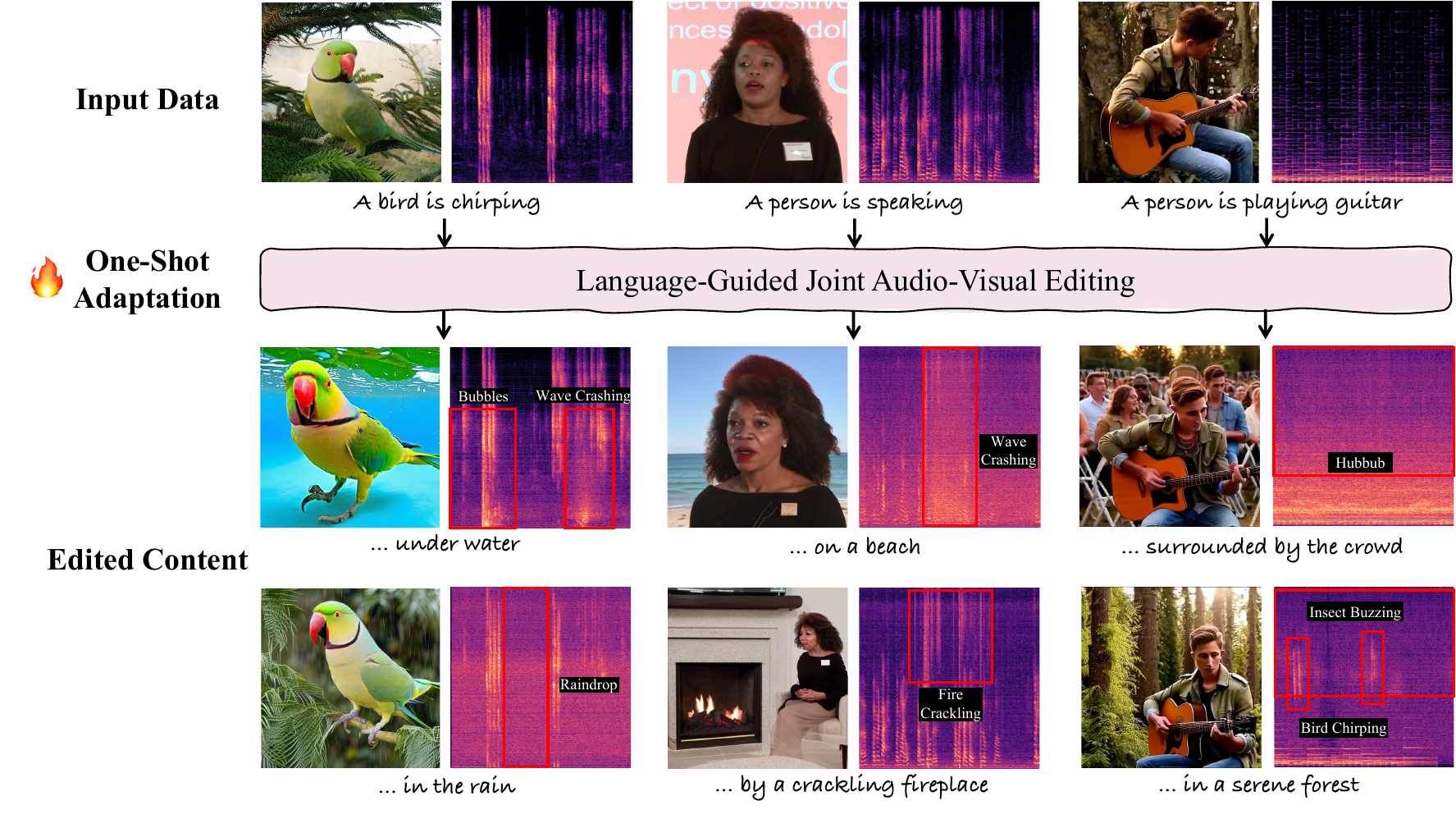}
    \captionof{figure}{We propose a novel language-guided joint audio-visual editing approach that allows users to edit their own sounding objects conditioned on various instructions. It only requires as few as one audio-visual pair for adaptation and enables the generation of new audio-visual instances based on creative text prompts. For example, after we update diffusion models with the user-provided ``\textit{a bird is chirping}'' data, we can easily generate the image of a bird chirping under water and synthesize the same chirping sound mixed with the audio of water bubbles and wave crashing, conditioned on the free-form text prompt: ``\textit{a bird is chirping under water.}'' }
    \label{fig:teaser}
\end{figure}

\input{sec/1_introduction}
\input{sec/2_relate}
\input{sec/3_method}
\input{sec/4_experiment}
\input{sec/5_conclusion}

%
%
\bibliographystyle{splncs04}
\bibliography{main}
\end{document}

%% file: sec/0_abstract.tex
\begin{abstract}
In this paper, we introduce a novel task called language-guided joint audio-visual editing. Given an audio and image pair of a sounding event, this task aims at generating new audio-visual content by editing the given sounding event conditioned on the language guidance. For instance, we can alter the background environment of a sounding object while keeping its appearance unchanged, or we can add new sounds contextualized to the visual content.
To address this task, we propose a new diffusion-based framework for joint audio-visual editing and introduce two key ideas. Firstly, we propose a one-shot adaptation approach to tailor generative diffusion models for audio-visual content editing. With as few as one audio-visual sample, we jointly transfer the audio and vision diffusion models to the target domain. After fine-tuning, our model enables consistent generation of this audio-visual sample. Secondly, we introduce a cross-modal semantic enhancement approach. We observe that when using language as content editing guidance, the vision branch may overlook editing requirements. This phenomenon, termed catastrophic neglect, hampers audio-visual alignment during content editing. We therefore enhance semantic consistency between language and vision to mitigate this issue.
Extensive experiments validate the effectiveness of our method in language-based audio-visual editing and highlight its superiority over several baseline approaches. We recommend that readers visit our project page for more details: \url{https://liangsusan-git.github.io/project/avedit/}.

\end{abstract}

%% file: sec/1_introduction.tex
\section{Introduction}
\label{sec:intro}
The perception of real-world sounds and visual objects depends on the environment in which they occur. Similarly, the experience driven by different digital audio and visual artifacts is also expected to be characterized by the context around them. When we see an accordion and hear its sound, can we contextualize this audio-visual phenomenon in a different environment? Or can we picture it being played on a rainy day, accompanied by the pitter-patter of raindrops? How would it sound in a large hall? As we move more and more towards AI-driven generation of multimodal content, generative approaches to enable such capabilities are desirable. 
We answer these questions by proposing a new task: language-guided joint audio-visual editing. For an object that emits sound, we collect an audio and image pair of this object as the reference for the sounding object. The goal of our task is to generate new audio-visual content by modifying the reference data as per the user's natural language guidance. As shown in Fig.~\ref{fig:teaser}, this task edits the user-provided specific \textit{bird} to generate new audio-visual samples based on different prompts, \textit{e.g.}, ``\textit{a bird is chirping under water}.''
This novel task enhances user experience by enabling multimedia content editing. To the best of our knowledge, this is \emph{the first work to study natural language-guided editing of audio-visual content.}

Considering the powerful generation capability and language controllability of generative diffusion models \cite{stablediffusion,imagen,audioldm,make-an-audio,make-a-video}, one can adopt existing generative diffusion models and repurpose them for the language-guided audio-visual editing task. However, utilizing these diffusion models for this novel task poses two obstacles.
(1) Audio-visual editing necessitates generative models to \textbf{jointly} replicate the image and sound of the audio-visual object while adhering to the user's guidelines for editing. Although text prompts can be utilized to steer generative models towards producing sounding objects that are similar to the user-provided data, it is non-trivial to guide these models to generate the same sounding object. This is especially difficult for the simultaneous production of audio and visuals. Thus, designing an effective adaptation approach is crucial for adjusting audio-visual generative models to audio-visual content editing. (2) When using language guidance to control the audio-visual editing, we expect that the edited sample should reflect the prompts both visually and acoustically. However, we observe that vision models, such as Stable Diffusion \cite{stablediffusion}, tend to ignore instructions (termed catastrophic neglect \cite{chefer2023attend}) and generate content with poor audio-visual semantic alignment, failing to meet user expectations.

To overcome these challenges, we propose a multimodal one-shot adaptation approach. This adaptation involves jointly transferring the audio and vision models to the domain of the sounding object. We extract a compact yet representative feature from the given audio-visual sample, capturing its unique and multimodal characteristics. This feature serves as a guide for fine-tuning the diffusion model. By incorporating this meaningful representation, we enable the diffusion models to learn and memorize the specific audio-visual sample. Following fine-tuning, the optimized audio-visual diffusion model is capable of editing the given event based on textual instructions while retaining the characteristics of the given audio-visual sample.

Moreover, we design a cross-modal semantic enhancement method to mitigate the issue of catastrophic neglect within the vision branch. This method aims to enhance the semantic correlation between language and vision, achieved by adjusting the weights of vision-language attention maps. By emphasizing the user requirements during the vision editing process, we ensure that the generated images faithfully adhere to the language cues. Consequently, our approach facilitates consistent audio-visual content editing under language guidance.

Considering that our work is the first to study the language-based audio-visual editing problem, there is no publicly available benchmark to assess the editing performance. Therefore, we collect a one-shot audio-visual editing (OAVE) dataset comprising 44 distinct sounding events for benchmarking purposes. This dataset encompasses a diverse range of audio-visual samples, including animals, vehicles, musical instruments, human speech, tools, and natural phenomena. We believe this dataset will lay the foundation for studying more general and more complex sounding objects. Extensive experiments on the OAVE dataset demonstrate that our approach can effectively edit audio-visual content conditioned on language guidance. Our framework shows promising editing performance from both subject fidelity and prompt faithfulness aspects.

%% file: sec/2_relate.tex
\section{Related Work}
Our work is related to one-shot content editing, audio-visual generation, and diffusion models. We discuss each of these topics in the following sub-sections.

\subsection{One-Shot Content Editing}
One-shot content editing based on generative models can be classified into two categories. The first category \cite{song2020denoising,dhariwal2021diffusion,prompt2prompt} achieves content editing without model fine-tuning. DDIM Inversion \cite{song2020denoising,dhariwal2021diffusion} inverses an input image into a latent noise by performing the diffusion process in reverse order. It edits the image by controlling the denoising process of this latent noise with a text condition. Prompt2Prompt \cite{prompt2prompt} conducts content editing based on the DDIM Inversion. It adjusts the cross-attention map to alter the effects of language on content editing. The second category \cite{text-inversion,dreambooth,custom-diffusion,animate-a-story,instantbooth,chen2023disenbooth,xiao2023fastcomposer,hua2023dreamtuner,wei2023elite} realizes content editing by fine-tuning the generative model with a one-shot sample. Textual Inversion~\cite{text-inversion} introduces an inversion approach to find in the embedding space a special token that enables Stable Diffusion to reconstruct the user-provided images. During optimization, the parameters of the diffusion model and the text encoder are frozen. DreamBooth~\cite{dreambooth} enhances the editing fidelity by fine-tuning the whole diffusion model. Custom Diffusion~\cite{custom-diffusion} is a compromise between Textual Inversion and DreamBooth, which only updates key and value matrices in cross-attention modules to improve training efficiency and avoid overfitting. Animate-A-Story~\cite{animate-a-story} further suggests using low-rank adaptation technique~\cite{lora} to achieve one-shot editing. While these methods yield satisfactory results for vision editing, their direct application in audio-visual editing often results in misaligned content due to the isolation of audio and vision learning.

\subsection{Audio-Visual Generation}
Audio-visual generation aims to synthesize audio and visual content by utilizing cross-modal correlation between the two modalities. Several recent studies~\cite{sound2scene,tpos,lee2022sound,gluegen} perform audio-to-image generation by aligning audio features with latent conditions for controllable image synthesis. Conversely, some works~\cite{make-an-audio,audioldm2,luo2023difffoley,du2023conditional,mo2023diffava,avnerf,liang2023neural,huang2023davis,huan2024modeling} focus on the vision-to-audio generation, extracting semantic and temporal information from visual input for audio synthesis. Meanwhile, MM-Diffusion~\cite{mm-diffusion} proposes jointly generating audio-visual content to enhance audio-visual relevance built upon unconditional diffusion models. CoDi~\cite{any-to-any} introduces a language-based modality ``bridging'', that projects different modalities into a joint multimodal space, enabling text-to-audio-visual generation. MUGEN~\cite{mugen} presents a unified auto-regressive transformer for audio-visual generation on a synthetic dataset. While these works achieve promising results, none have effectively addressed language-based audio-visual editing. MM-Diffusion does not consider generation conditions, resulting in random audio-visual content. The language conditions used by CoDi and MUGEN provide only coarse and limited control over content creation, failing to meet specific user requirements.

\subsection{Diffusion Models}
Diffusion models have shown impressive success in text-to-image~\cite{GLIDE,dalle2,imagen,stablediffusion}, text-to-video~\cite{ho2022video,make-a-video,tune-a-video,text2video-zero,animatediff}, and text-to-audio tasks~\cite{diffsound,audioldm,make-an-audio,tango,audioldm2,make-an-audio2}. GLIDE~\cite{GLIDE} proposes text-conditional diffusion models for controllable image synthesis. DALL-E 2~\cite{dalle2} introduces a two-stage diffusion model that utilizes the joint CLIP representation~\cite{clip} for textual condition extraction. Imagen~\cite{imagen} discovers that a pretrained large language model (e.g., T5~\cite{t5}) can effectively encode text prompts for image generation. Latent Diffusion Model (Stable Diffusion)~\cite{stablediffusion} suggests conducting diffusion and denoising processes in the latent space to improve training efficiency while retaining generation quality. Several works, such as DiffSound~\cite{diffsound}, AudioLDM~\cite{audioldm}, Make-An-Audio~\cite{make-an-audio}, and TANGO~\cite{tango}, extend diffusion models to the audio domain. Even though these methods exhibit promising generation capabilities, they face challenges in accommodating users' specific editing demands.

%% file: sec/3_method.tex
\begin{figure}[t]
    \centering
    \includegraphics[width=\linewidth]{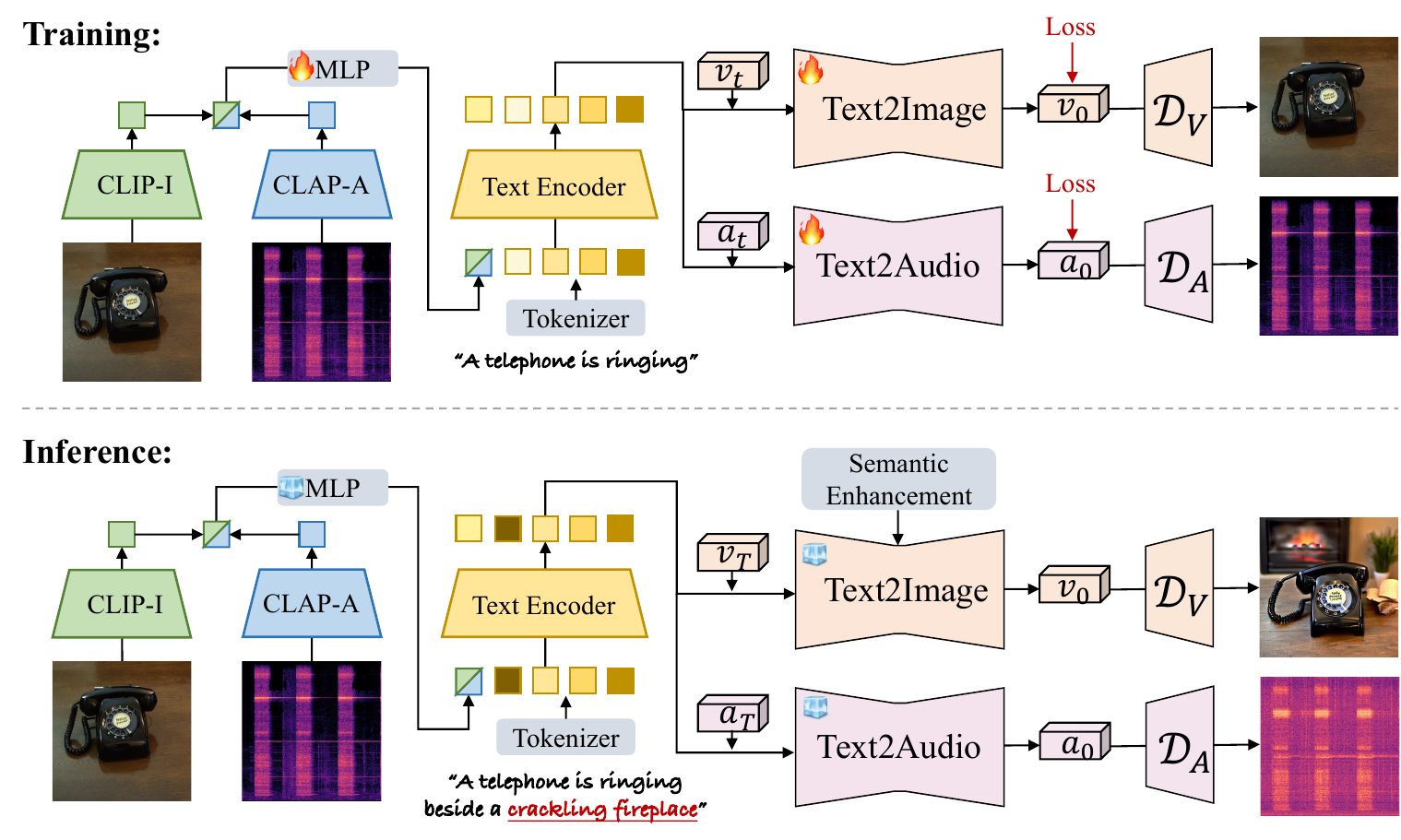}
    \caption{Our framework for language-guided audio-visual editing. During \textbf{training}, we extract unimodal information from the audio-visual sample using pretrained encoders. Then, we fuse audio and visual features with an MLP and feed the output along with the text prompt into the text encoder. The text encoder generates textual conditions to guide the audio-visual diffusion model. We update the parameters of the MLP and diffusion models. During \textbf{inference}, we freeze all parameters of our model. We replace the training prompt with an editing prompt, e.g., we append ``\textit{beside a crackling fireplace}''  to the training prompt ``\textit{a telephone is raining}.'' We inject the cross-model semantic enhancement module into the vision branch to improve semantic consistency. The generated audio and image accurately reflect the editing requirements.}
    \label{fig:framework}
\end{figure}

\section{Method}
We propose a diffusion model-based content editing framework to solve the language-guided joint audio-visual editing task. We first introduce preliminary knowledge about diffusion models in Sec.~\ref{sec:background}. Our audio-visual diffusion model is detailed in Sec.~\ref{sec:avdiff}. Then, we illustrate our multimodal one-shot adaptation approach in Sec.~\ref{sec:condition}, which jointly adjusts the audio-visual generative model to content editing. Moreover, we present a cross-modal semantic enhancement method to emphasize the targeted editing effects (Sec.~\ref{sec:enhance}). In Fig.~\ref{fig:framework}, we provide an overview of our language-guided audio-visual editing framework, including both training and inference pipelines.

\subsection{Background}
\label{sec:background}

\textbf{Latent diffusion models.} Latent Diffusion Model~\cite{stablediffusion} conducts diffusion and denoising processes in the latent space to improve the training efficiency. A pair of autoencoders $\mathcal{E}$ and $\mathcal{D}$, implemented as VQ-GAN~\cite{esser2021taming} or VQ-VAE~\cite{van2017neural}, are used to perform dimension reduction and dimension recovery. Specifically, given an input $X\in\mathbb{R}^{C\times H \times W}$, where $C$ is the number of channels and $H$ and $W$ are the spatial dimensions of the input, the latent diffusion model firstly reduces its spatial dimension to $x_0=\mathcal{E}(X)\in \mathbb{R}^{c\times h \times w}$, where $\frac{H}{h}=\frac{W}{w}$ are the downsampling scale and $c$ is the number of channels of $x_0$. Subsequently, the diffusion process iteratively injects Gaussian noise into the latent feature $x_0$ to destroy it:
\begin{equation}
    \label{eq:diffusion}
    q(x_t|x_{t-1})=\mathcal{N}(x_t; \sqrt{1 - \beta_t}x_{t-1}, \beta_t I), t=1,\dots,T,
\end{equation}
where $\{\beta_t\}_{t=1}^T$ are hyper-parameters to control the strength of injected noise. $T$ represents the number of diffusion steps and is usually set to $1000$. After noise injection, the latent feature $x_T$ is expected to follow $\mathcal{N}(0,I)$. The denoising process iteratively eliminates injected noise from the latent feature $x_t$ to recover the initial signal $x_0$:
\begin{equation}
    \label{eq:denoise}
    p(x_{t-1}|x_t)=\mathcal{N}(x_{t-1}; \mu_\theta(x_t,t), \Sigma_\theta(x_t,t)), t=1,\dots,T,
\end{equation}
where $\mu_\theta$ and $\Sigma_\theta$ are trainable models to estimate the mean and variance of the distribution. Once the signal $x_0$ is recovered, the decoder $\mathcal{D}$ projects $x_0$ to the original space $X$.

Following DDPM~\cite{ddpm}, we set $\Sigma_\theta$ to untrained constants. We use the following objective to train a latent diffusion model: 
\begin{equation}
\begin{split}
    &\mathbb{E}_{x,\epsilon\sim\mathcal{N}(0,1),t}\left[ ||\epsilon-\epsilon_\theta (x_t, t)||_2^2 \right],
\end{split}
\label{eq:ddpm}
\end{equation}
where $x_t=\sqrt{\bar{\alpha_t}}x_0 + \sqrt{1-\bar{\alpha_t}}\epsilon$, $\bar{\alpha_t}=\prod_{i=1}^{t}\alpha_i$, and $\alpha_t=1-\beta_t$. $\epsilon_\theta$ is a learnable diffusion model and commonly implemented as a U-Net network~\cite{unet}.

\noindent \textbf{Language-guided diffusion models.} To enable the diffusion model for text-conditioned generation, a text encoder $\tau$ (\textit{e.g.}, CLIP~\cite{clip} or CLAP~\cite{clap} text encoder) is used for extracting textual embedding from the input prompt $y$. The textual embedding $\tau(y)$ is then fed into the U-Net $\epsilon_\theta$ to control the denoising process via the cross-attention mechanism. The training objective in Eq.~\ref{eq:ddpm} is updated with
\begin{equation}
    \label{eq:loss_ldm}
    \mathbb{E}_{x,\epsilon\sim\mathcal{N}(0,1),t}\left[ ||\epsilon-\epsilon_\theta (x_t, \tau(y), t)||_2^2 \right].
\end{equation}

\subsection{Language-Guided Audio-Visual Generative Model}
\label{sec:avdiff}
To accomplish language-guided joint audio-visual editing, we first develop a text-conditioned audio-visual generative model. Instead of training a text-to-audio-visual diffusion model from scratch, we leverage pretrained unimodal diffusion models, namely Stable Diffusion~\cite{stablediffusion} and AudioLDM 2~\cite{audioldm2}. These unimodal diffusion models have been trained on extensive datasets and demonstrate robust generation capabilities, facilitating content synthesis guided by language.

We begin with an audio-visual pair $(A,V)$, where $A\in \mathbb{R}^{L}$ represents the waveform audio of length $L$, and $V\in \mathbb{R}^{C \times H \times W}$ represents an image with dimensions $C,H,$ and $W$. We convert the audio $A$ to the mel-spectrogram $A'\in\mathbb{R}^{1\times H' \times W'}$ using STFT, where $H'$ and $W'$ are the time-frequency sizes of $A'$. We utilize the pretrained autoencoder $\mathcal{E}_A$ from AudioLDM 2 \cite{audioldm2} to project $A'$ into the latent space with $a_0=\mathcal{E}_A(A')\in \mathbb{R}^{c_A \times h_A \times w_A}$. Similarly, we use the pretrained autoencoder of Stable Diffusion \cite{stablediffusion} $\mathcal{E}_V$ to map $V$ to a latent feature $v_0=\mathcal{E}_V(V)\in \mathbb{R}^{c_V \times h_V \times w_V}$. Then we sample a diffusion step $t\in[1,1000]$ and use Eq.~\ref{eq:diffusion} to generate altered samples $a_t$ and $v_t$.

Given a text prompt $y$, we adopt the CLIP text model $\tau_V$ \cite{clip} to extract vision-related textual conditions from $y$, and the CLAP text model $\tau_A$ \cite{clap} to obtain audio-sensitive conditions. The extracted conditions $\tau_A(y)$ and $\tau_V(y)$ are used to guide the denoising process of the audio-visual diffusion model, i.e., $\epsilon_{\theta_A} (a_t, \tau_A(y), t)$ and $\epsilon_{\theta_V} (v_t, \tau_V(y), t)$. $\epsilon_{\theta_A}$ and $\epsilon_{\theta_V}$ are U-Net networks introduced in \cite{audioldm2} and \cite{stablediffusion}, respectively. Once the $a_0$ and $v_0$ are generated, we use the pretrained decoder $\mathcal{D}_V$ to recover the spatial dimension of $v_0$. Similarly, we use $\mathcal{D}_A$ to decode $a_0$ to $A'$. We utilize the HiFi-GAN vocoder~\cite{kong2020hifi} to recover the waveform audio $A$ from the reconstructed mel-spectrogram $A'$. Consequently, we can synthesize an audio-visual pair based on the input prompt.

\subsection{Multimodal One-Shot Adaptation}
\label{sec:condition}

\begin{wrapfigure}[17]{r}{0.5\textwidth}
    \centering
    \includegraphics[width=\linewidth]{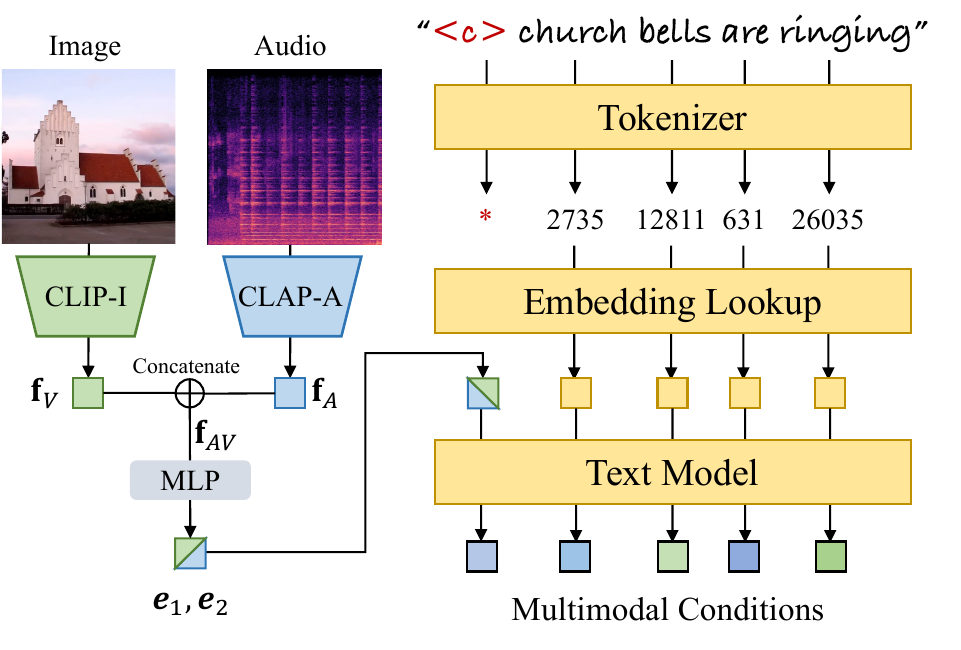}
    \caption{Multimodal one-shot adaptation. We extract meaningful audio-visual representations from user-provided data. We incorporate representations into textual embeddings and feed them to the text model to generate multimodal conditions. }
    \label{fig:condition}
\end{wrapfigure}

We aim to adapt the text-conditioned audio-visual generative model to address the language-guided joint audio-visual editing problem. Audio-visual editing necessitates the generative model to learn and memorize the audio-visual content of a given sample. However, generative models face challenges in accurately reproducing specific audio-visual content solely based on textual cues. To tackle this limitation, we propose a multimodal one-shot adaptation approach that transfers generative models to the target domain of this particular audio-visual sample. We illustrate our approach in Fig.~\ref{fig:condition} and the training part of Fig.~\ref{fig:framework}.

Specifically, given the same audio-visual pair $(A, V)$, we utilize the pretrained CLIP image encoder to extract a compact visual feature $f_V \in \mathbb{R}^d$ from $V$ and use the pretrained CLAP audio encoder to convert $A$ to a latent audio feature $f_A \in \mathbb{R}^d$, where $d$ is the dimension of feature vectors. We concatenate $f_V$ and $f_A$ as an audio-visual feature $f_{AV}$ to represent the multimodal characteristics of the sounding event. Since the diffusion model is controlled by the language condition, we convert the audio-visual feature $f_{AV}$ to text-compatible representations using Multi-Layer Perceptrons (MLPs). We design two MLPs to project the $f_{AV}$ to the embedding spaces of $\tau_A$ and $\tau_V$, respectively.
\begin{equation}
    e_1 = \mathrm{MLP}_1(f_{AV}),  e_2 = \mathrm{MLP}_2(f_{AV}).
    \label{eq:mlp}
\end{equation}

For a training prompt $y$, we add a placeholder token \texttt{<c>} before the tokens of the sounding object that we aim to memorize. For instance, in the prompt ``\textit{church bells are ringing},'' we insert \texttt{<c>} before ``\textit{church bells}.'' Next, we tokenize the modified prompt by mapping each word to a unique dictionary id. We generate an embedding sequence by looking up the embedding of each token according to its id. Since \texttt{<c>} is a placeholder with no corresponding embedding, we use the text-compatible representations $e_1,e_2$ as its embedding. The embedding sequence is then fed into the text model to encode conditions with rich multimodal information about the sounding object. We feed the embedding sequence with $e_1$ into $\tau_A$, and the embedding sequence with $e_2$ into $\tau_V$. To maintain clarity, we depict only one text model in Fig.~\ref{fig:framework} and Fig.~\ref{fig:condition}.

We leverage these multimodal conditions to optimize the audio-visual diffusion model to learn and memorize the particular sounding object. We use the following objective to accomplish multimodal one-shot adaptation:
\begin{equation}
    \label{eq:loss_av}
    \begin{split}
    &\mathbb{E}_{a,\epsilon_A\sim\mathcal{N}(0,1),t}\left[ ||\epsilon_A-\epsilon_{\theta_A} (a_t, \tau_A(e_1,y), t)||_2^2 \right] + \\
    & \mathbb{E}_{v,\epsilon_V\sim\mathcal{N}(0,1),t}\left[ ||\epsilon_V-\epsilon_{\theta_V} (v_t, \tau_V(e_2,y), t)||_2^2 \right].
    \end{split}
\end{equation}
Once we finish the one-shot adaptation, the fine-tuned model is capable of generating audio-visual pairs that closely resemble the given audio-visual sample.

\subsection{Cross-Modal Semantic Enhancement}
\label{sec:enhance}
After we adjust the audio-visual model, we can utilize it for language-guided joint audio-visual editing. By replacing the training prompt $y$ with a new editing prompt $y^*$, we can edit the user-provided sounding object and generate new audio-visual content. For example, we alter the environment of the sounding event or we add another sounding object into the audio-visual event.

\begin{figure}[t]
    \centering
    \includegraphics[width=\textwidth]{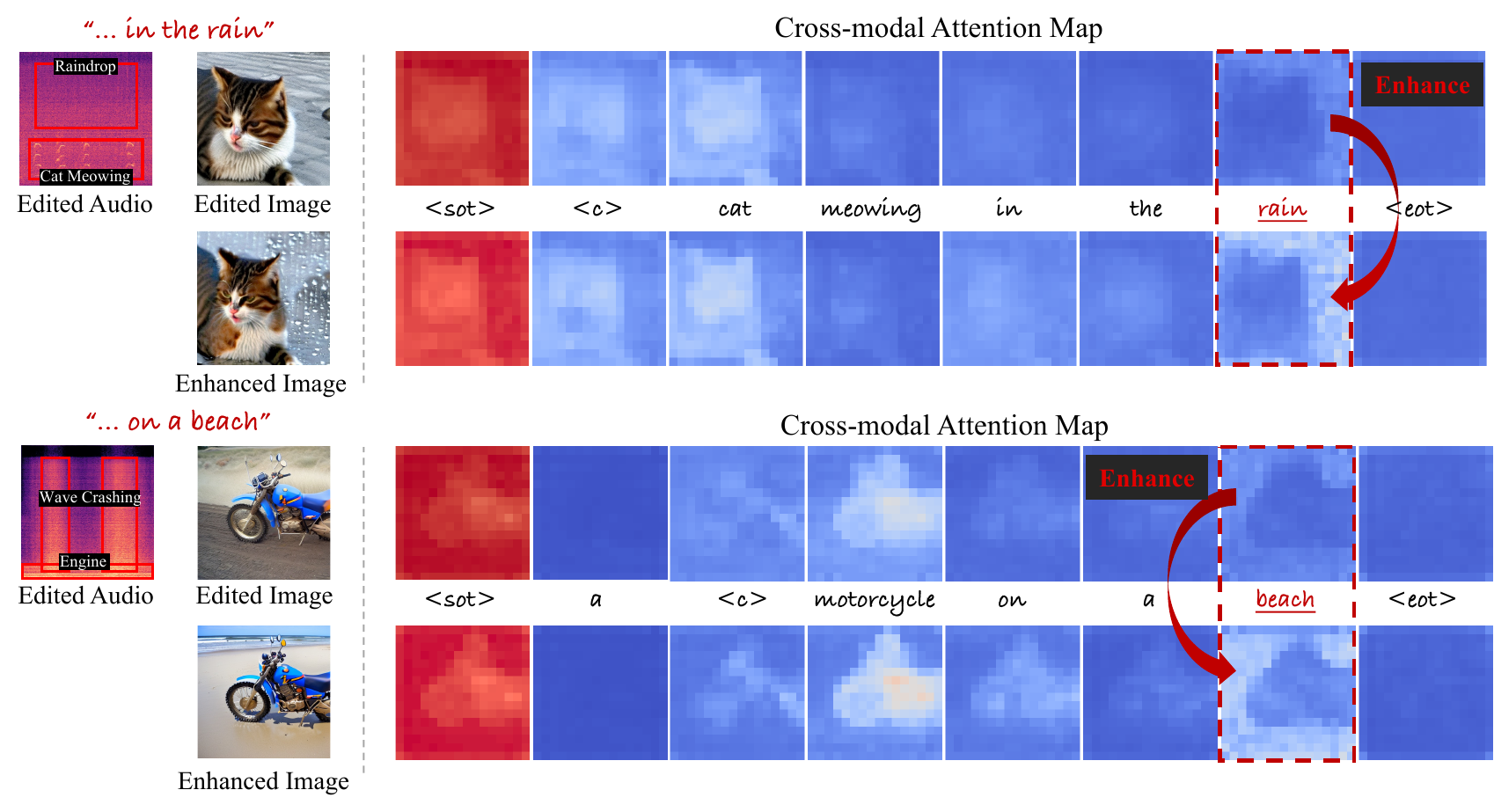}
    \caption{Cross-modal semantic enhancement. The vision model tends to neglect the editing requirements while the audio model can accurately generate targeted content. We adjust the weights of vision-language attention maps to mitigate this issue. Eventually, we achieve consistent audio-visual content editing conditioned on language.}
    \label{fig:neglect}
\end{figure}

However, we observe that the vision branch tends to ignore some editing requirements specified by the text prompt. In Fig.~\ref{fig:neglect}, we present examples generated by our model to highlight this issue. While the audio branch is capable of synthesizing the sound of raindrops to complement the editing requirement ``\textit{in the rain}'' and can include the sound of waves crashing to match ``\textit{on a beach},'' the vision branch fails to produce the corresponding visual elements. This phenomenon, termed ``catastrophic neglect'' \cite{chefer2023attend}, leads to inconsistent audio-visual editing outcomes, consequently reducing overall user satisfaction.

To address this limitation, we propose a cross-modal semantic enhancement approach. Our approach is built upon the following analysis. When we use a text prompt $y^*$ to edit the visual latent feature $v_t$, we project $v_t$ to a query matrix, and convert textual conditions $\tau_V(e_2,y^*)$ into key and value matrices. A cross-modal attention map $M$ is calculated between the image query and the text key, representing the contribution of each token to each image patch. When employing a prompt $y^*$ for content editing, we expect that all meaningful tokens, rather than just a subset, should have certain attention weights to guide the image generation process. However, as shown in Fig.~\ref{fig:neglect}, we notice that (1) the ignored tokens have low cross-attention weights, which means these tokens have limited influence over image editing; (2) a special token \texttt{<sot>}, marking the beginning of the text, is assigned significantly higher cross-attention weight compared to other tokens, thereby dominating the cross-modal guidance.

To enhance the semantic correlation between language and vision, we propose adjusting the cross-attention map $M$, where $M_{i,j}\in \mathbb{R}$ indicates the influence of the $j$-th token on the $i$-th image patch. We decrease the importance of the token \texttt{<sot>} and emphasize the editing requirements (tokens that are in the editing prompt $y^*$ but not in the training prompt $y$). For a token sequence $P=\{p_1, p_2, \dots, p_N\}$, we scale the cross-modal attention weights as follows:
\begin{equation}
    M_{i,j}^*:=     \begin{cases}
      \alpha \cdot M_{i,j} &\quad\text{if }p_j=\texttt{<sot>}, \\
      \beta \cdot M_{i,j} &\quad\text{if }p_j\in y^*\text{and } p_j \notin y, \\
      M_{i,j} &\quad\text{otherwise.} \\ 
     \end{cases}
\end{equation}
We set $\alpha\in[0,1]$ to reduce the impact of the \texttt{<sot>} token and $\beta \in [1,4] $ to emphasize the editing requirements. By enhancing the semantic correlation between vision and language, we attain consistent audio-visual content editing.

%% file: sec/4_experiment.tex
\section{Experiment}
\subsection{Dataset and Setup}

\begin{figure}[t]
    \centering
    \includegraphics[width=\linewidth]{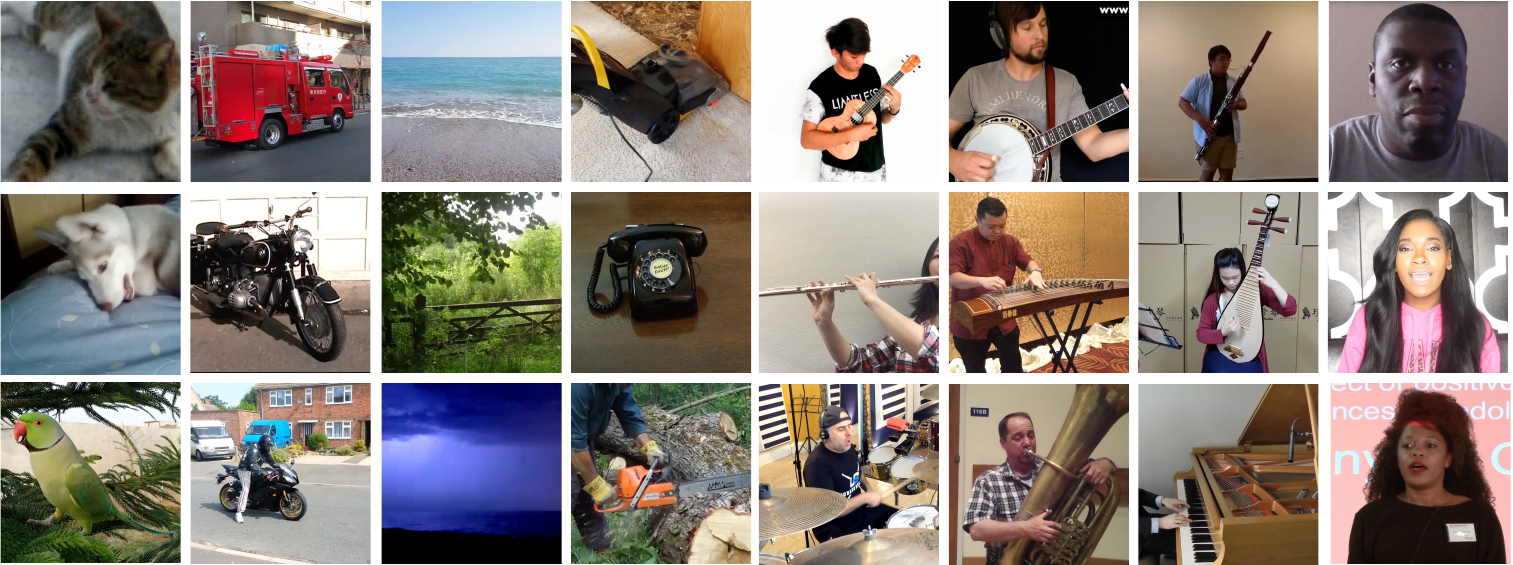}
    \caption{We show some samples from the OAVE dataset, including animals, vehicles, tools, natural phenomena, musical instruments, and human speech.}
    \label{fig:dataset}
\end{figure}

\noindent \textbf{Datasets.} We curate the One-shot Audio-Visual Editing (OAVE) dataset to evaluate different language-guided joint audio-visual editing approaches. The OAVE dataset consists of 44 distinct sounding events for benchmarking purposes, encompassing animals, vehicles, tools, natural phenomena, musical instruments, and human speech. We collect data from the MUSIC~\cite{zhao2018sound}, AVSpeech~\cite{avspeech}, and VGGSound \cite{vggsound} datasets. Fig.~\ref{fig:dataset} presents some examples from the OVAE dataset. This dataset lays a foundation for future work on more complex and diverse sounding objects. We will release this dataset to the research community.

For each audio-visual sample, the dataset provides one video, from which we extract one 10-second audio clip and 10 corresponding frames. We collect 25 prompts for language-guided joint audio-visual editing. These prompts either inject new sounding objects (e.g., dog, train, child) into the user-provided content or adjust the environment (e.g., cathedral, forest, underwater) of the given audio-visual event, influencing the visual context and acoustic property.

\noindent \textbf{Evaluation Metrics.} We evaluate the performance of language-based audio-visual editing from \textit{subject fidelity} and \textit{prompt faithfulness} perspectives. From the first aspect, we design metrics to measure the similarity between training samples and the content synthesized by generative models, indicating whether generative models can replicate the user-provided content after adaptation. Specifically, we use CLIP-I and DINO metrics to assess the image similarity, following the approach in \cite{dreambooth}. CLIP-I metric calculates the pairwise similarity between reference and generated images using CLIP image encoder~\cite{clip}. DINO metric computes the pairwise similarity with self-supervised DINO ViT~\cite{dino}. Similarly, we adopt CLAP-A and FAD \cite{DBLP:conf/interspeech/KilgourZRS19} to measure the audio similarity.

From the second perspective, we utilize CLIP-T, CLAP-T, and AVSS to measure the prompt faithfulness of generated content. CLIP-T works similarly to CLIP-I but calculates the similarity between generated images and text prompts. CLAP-T calculates the audio-text similarity. For measuring semantic consistency between generated audio and image pairs, we define a new metric called Audio-Visual Semantic Similarity (AVSS). AVSS determines the semantic consistency between generated audio and image using the AudioCLIP model~\cite{audioclip}.

\begin{table*}[t]
    \centering
    \caption{Comparison with existing one-shot language-guided content editing approaches. We reproduce these methods on our OAVE dataset and report their performance results. We design various metrics to measure the editing quality from subject fidelity and prompt faithfulness aspects. The method ``Real'' represents performance results evaluated using ground-truth samples, serving as an upper bound for this task. }
    \resizebox{\linewidth}{!}{
    \begin{tabular}{c|cccc|ccc}
    \toprule
       Methods  & CLIP-I $\uparrow$ & DINO $\uparrow$ & CLAP-A $\uparrow$ & FAD $\downarrow$ & CLIP-T $\uparrow$  & CLAP-T $\uparrow$  & AVSS $\uparrow$ \\
    \midrule
       Real & 0.952 & 0.937 & 0.948 & -- & --  & -- & 0.139 \\
       \midrule
       Animate-A-Story \cite{animate-a-story} & 0.578 & 0.385 & 0.526 & 16.139 & 0.626 & 0.157 & 0.053\\
       Textual Inversion \cite{text-inversion} & 0.714 & 0.574 & 0.300 & 38.049 & 0.788 & 0.150  & 0.068\\
       Custom Diffusion \cite{custom-diffusion} & 0.788 & 0.699 & 0.538 & 17.172 & 0.664  & 0.154 & 0.118 \\
       DreamBooth \cite{dreambooth} & 0.662 & 0.576 & 0.530 & 15.927 & \textbf{0.824}  & 0.209 &  0.108\\
       \midrule
       Ours & \textbf{0.823} & \textbf{0.737} & \textbf{0.593} & \textbf{15.541} & 0.788  & \textbf{0.228} & \textbf{0.120} \\
    \bottomrule
    \end{tabular}
    }
    \label{tab:sota}
\end{table*}

\noindent\textbf{Implementation Details.}
We select AudioLDM 2 as the audio diffusion model and Stable Diffusion v1.5 as the image diffusion model. The two MLPs used in Eq.~\ref{eq:mlp} have the same structure. Each MLP consists of two linear layers with a width of 1024, and a ReLU activation layer is applied between linear layers. When conducting multimodal one-shot adaptation, we set the learning rate of the audio branch as $5e{-}5$, the vision branch as $5e{-}5$, and the MLP as $1e{-}4$. We use the Adam optimizer to optimize parameters, performing $300$ steps with a batch size of 1. We use the DDPM scheduler~\cite{ddpm} (1000 diffusion steps) for training and the DDIM scheduler~\cite{ddim} (50 steps) for inference. 

\noindent \textbf{Baselines.} To the best of our knowledge, this is the first work to tackle the language-guided joint audio-visual editing task via one-shot adaptation. Thus, we adopt existing one-shot text-based vision editing approaches as baselines, including Textual Inversion~\cite{text-inversion}, DreamBooth~\cite{dreambooth}, Custom Diffusion~\cite{custom-diffusion}, and Animate-A-Story~\cite{animate-a-story}. We extend these methods for this new task by applying the proposed approaches to both audio and vision branches. The audio and vision models learn the modality-specific features of the sounding object individually.

\subsection{Quantitative Evaluation}
\noindent \textbf{Comparison with existing approaches.} We report one-shot language-guided joint audio-visual editing results of different methods in Tab.~\ref{tab:sota}. The method ``Real'' denotes the performance measured between ground-truth samples, meaning an upper bound of this task. Our method surpasses other baselines, achieving $0.823$ in the CLIP-I metric and $0.737$ in the DINO metric, showing our approach can accurately replicate the visual details. Furthermore, our method scores $0.593$ in the CLAP-A metric and $15.541$ in the FAD metric, highlighting its robustness in retaining auditory features. Regarding prompt faithfulness, our method attains $0.788$ CLIP-T similarity and $0.228$ CLAP-T similarity. Our method generates audio-visual pairs with the highest audio-visual semantic similarity score of $0.120$, demonstrating consistent audio-visual content editing through cross-modal semantic enhancement. DreamBooth achieves superior performance in the CLIP-T metric, possibly due to its fixed text model during finetuning.
\begin{table}[t]
    \centering
    \caption{Ablation studies. We analyze the influence of different design choices on language-guided joint audio-visual editing.}
    \resizebox{\linewidth}{!}{
    \begin{tabular}{cc|cccc|ccc}
    \toprule
    \multicolumn{2}{c|}{Methods}                                        & CLIP-I $\uparrow$ & DINO $\uparrow$ & CLAP-A $\uparrow$ & FAD $\downarrow$ & CLIP-T $\uparrow$  & CLAP-T $\uparrow$ & AVSS $\uparrow$ \\
    \midrule
    \multicolumn{1}{c|}{\multirow{3}{*}{Adaptation}}      & Text       & 0.701 & 0.545 & 0.529 & 20.741 & 0.814  & 0.231 & 0.110 \\
    \multicolumn{1}{c|}{}                                & Unimodal   &  0.775 & 0.636 & 0.529 & 21.673 & 0.784  & 0.233 & 0.118\\
    \multicolumn{1}{c|}{}                                & Multimodal & 0.823 & 0.737 & 0.593 & 15.541 & 0.788  & 0.228 & 0.120\\
    \midrule
    \multicolumn{1}{c|}{\multirow{2}{*}{Feature Fusion}} & Early      &  0.823 & 0.737 & 0.593 & 15.541 & 0.788  & 0.228 & 0.120\\
    \multicolumn{1}{l|}{}                                & Late       & 0.848 & 0.758 & 0.598 & 16.921 & 0.750  & 0.232 & 0.120\\
    \midrule
    \multicolumn{1}{c|}{\multirow{4}{*}{Enhancement}}      & $\alpha=0.4,\beta=4.0$ & 0.712 & 0.532 & -- & -- & 0.790  & -- & 0.110\\
    \multicolumn{1}{l|}{}                                & $\alpha=0.6,\beta=3.0$      & 0.823 & 0.737 & -- & -- & 0.788  & -- & 0.120\\
    \multicolumn{1}{l|}{}                                & $\alpha=0.8,\beta=2.0$     & 0.829 & 0.743 & -- & -- & 0.778  & -- & 0.124\\
    \multicolumn{1}{l|}{}                                & $\alpha=1.0,\beta=1.0$     & 0.766 & 0.618 & -- & -- & 0.756  & -- & 0.111 \\
    \bottomrule
    \end{tabular}
    }
    \label{tab:ablation}
\end{table}

\noindent \textbf{Ablation Studies.} We conduct the following ablation studies (see Tab.~\ref{tab:ablation}) to analyze the influence of each component of our model on language-guided joint audio-visual editing. We first examine the effects of different adaptation approaches, including text-only finetuning, unimodal finetuning, and multimodal adaptation (our proposed approach). The text-only adaptation approach \cite{text-inversion,dreambooth,custom-diffusion,animate-a-story} uses textual information to finetune diffusion models. The unimodal adaptation method \cite{wei2023elite,chen2023disenbooth,xiao2023fastcomposer} utilizes modality-specific information to update the corresponding model, e.g., updating the image model with the visual information and the audio model with the audio data. As shown in the table, our multimodal approach effectively memorizes user-provided audio-visual samples and retains attributes of these samples during content editing. While our multimodal adaptation approach causes a marginal performance decline in CLIP-T and CLAP-T metrics, it notably enhances subject fidelity and audio-visual semantic similarity.

Moreover, we explore the impact of different feature fusion methods when conducting multimodal one-shot adaptation. Early fusion involves feeding audio-visual features, denoted as $\mathbf{e}_1$ and $\mathbf{e}_2$, directly into the text embedding space, which serves as the input of the language model. Conversely, late fusion entails feeding these features into the text feature space, which is the output of the language model. We empirically find that both early fusion and late fusion yield similar results. However, due to the higher CLIP-T score achieved with early fusion, we choose it as our default feature fusion approach.

Finally, we investigate the optimal hyperparameters, $\alpha$ and $\beta$, for cross-modal semantic enhancement. A low $\alpha$ value diminishes the influence of the token \texttt{<sot>} on both image layout and semantics. A high $\beta$ score amplifies the impact of editing requests. With $\alpha=1.0$ and $\beta=1.0$, cross-modal semantic enhancement is disabled. Since this enhancement primarily affects the vision branch, we present all vision-related metrics in the table. Our findings are (1) increasing $\alpha$ and decreasing $\beta$ effectively enhance prompt faithfulness and audio-visual semantic consistency; (2) however, setting $\alpha=0.4$ and $\beta=4.0$ diminishes subject fidelity as the semantic information embedded in \texttt{<sot>} is nearly eliminated; (3) both $\alpha=0.6$, $\beta=3.0$ and $\alpha=0.8$, $\beta=2.0$ strike a similar trade-off. We opt for $\alpha=0.6$ and $\beta=3.0$ as our default hyperparameters.

\subsection{Qualitative Evaluation}
\begin{figure}[t]
    \centering
    \includegraphics[width=0.9\linewidth]{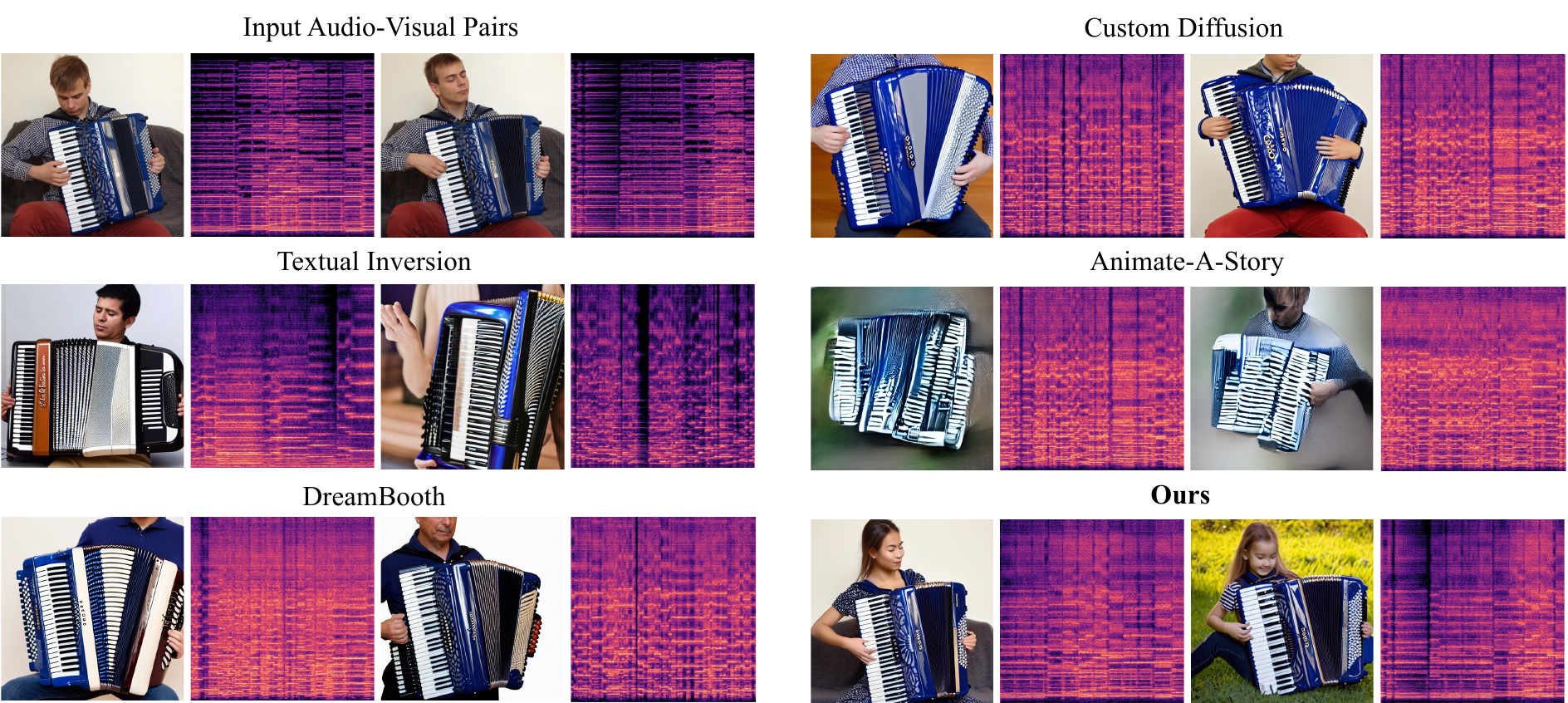}
    \caption{Visualization of edited audio-visual content. For each baseline approach, we show generated images and spectrograms with the prompt, ``\textit{A person plays accordion}.'' We challenge our method with two more complex prompts, ``\textit{A woman plays accordion}'' and ``\textit{A girl plays accordion on the grass}.'' for content editing. The original training data is displayed in the top-left corner for reference.}
    \label{fig:result}
\end{figure}

\noindent \textbf{Comparison with existing approaches.} We present visualization results in Fig.~\ref{fig:result} to intuitively evaluate performance. For each baseline approach, we show two pairs of audio-visual data edited using the prompt ``\textit{A person plays accordion}.'' To evaluate our model, we select two more challenging prompts, ``\textit{A woman plays accordion}'' and ``\textit{A girl plays accordion on the grass}.'' We include the input training data in the top-left corner. As shown in the figure, (1) our approach accurately reproduces the sounding object with the highest fidelity. The reconstructed accordion resembles the accordion used for training, capturing its dedicated appearance and texture. The generated audio shares the same melody as the training data. (2) Our approach enables editing instances and environments while preserving the details of the audio-visual sample. In the illustrated figures, our approach successfully replaces the person who plays accordion with a woman and a girl, and we adjust the environment from indoor to outdoor.

\begin{figure}[t]
\begin{subfigure}{0.5\textwidth}
  \centering
  \includegraphics[width=.9\linewidth]{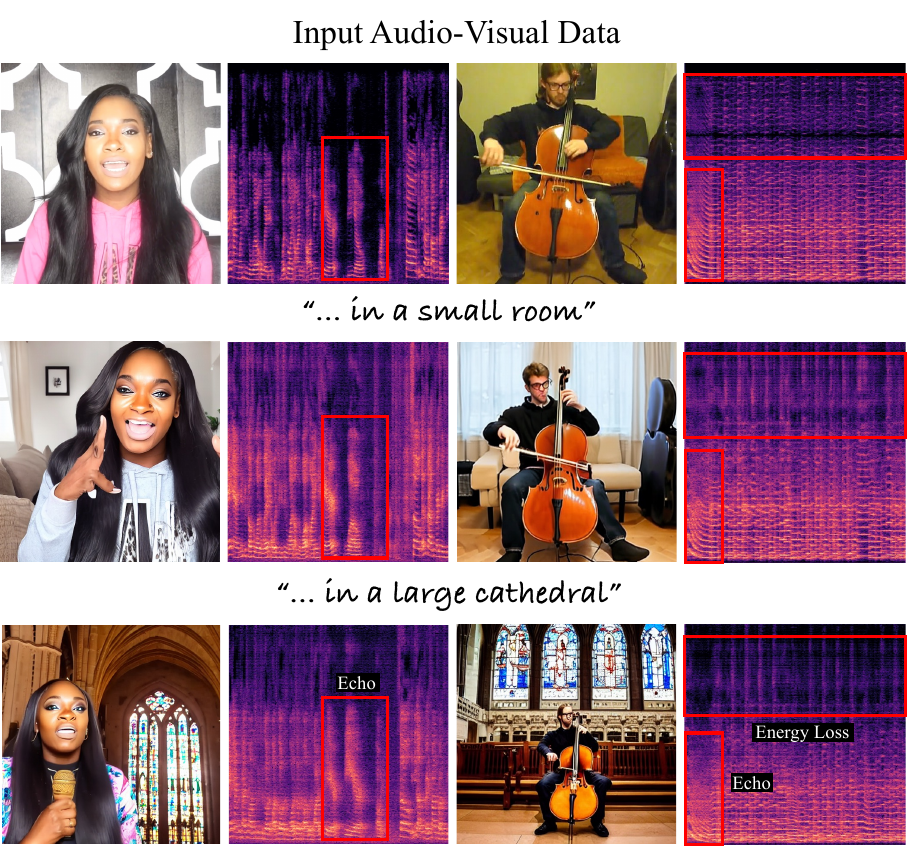}
  \caption{Room Acoustic Adjustment.}
  \label{fig:acoustic}
\end{subfigure}
\begin{subfigure}{0.5\textwidth}
  \centering
  \includegraphics[width=0.9\linewidth]{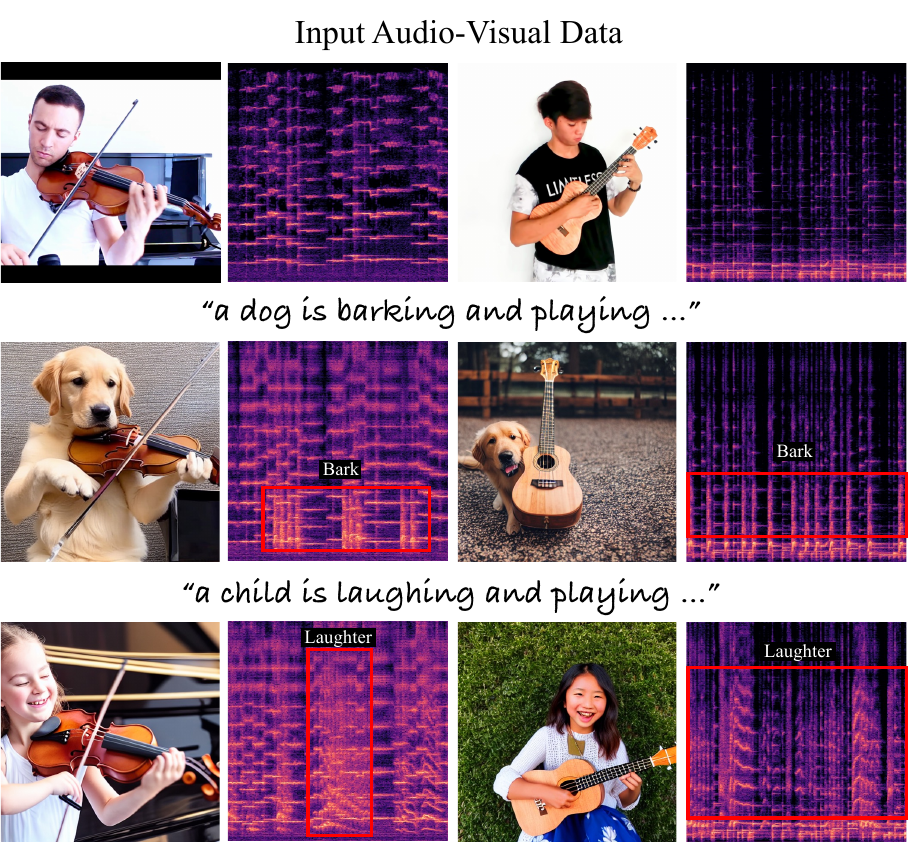}
  \caption{Audio-Visual Composition.}
  \label{fig:compose}
\end{subfigure}
\caption{(a) We place audio-visual samples in new environments and adjust the samples' visual and acoustic properties jointly. (b) We compose user-provided audio-visual samples with common sounding objects.}
\end{figure}

\noindent \textbf{Room Acoustic Adjustment.} Using our language-guided joint audio-visual editing framework, we can place sounding objects in new environments. We not only visually render the appearance and structure of an environment but also accordingly adjust the sound effects to match the audio with the room acoustics. In Fig.~\ref{fig:acoustic}, we place the sounding objects ``\textit{in a small room}'' and ``\textit{in a large cathedral}.'' Spectrograms produced in large cathedrals demonstrate noticeable reverberation and energy loss compared to those generated in small rooms, aligning with the acoustic characteristics of cathedrals.

\noindent \textbf{Audio-Visual Composition.} Our model can compose the user-provided audio-visual sample with common sounding objects. In Fig.~\ref{fig:compose}, we merge sounding objects with ``\textit{dog}'' and ``\textit{child}.'' This results in an image depicting a dog playing the violin, accompanied by an audio clip of violin music and barking sounds. Additionally, we synthesize a photo of a joyful child playing the ukulele and laughing, with a corresponding audio track blending ukulele music with laughter.

%% file: sec/5_conclusion.tex
\section{Conclusion}
This paper investigates the novel language-guided joint audio-visual editing problem and proposes a new diffusion-based editing framework. We incorporate multimodal one-shot adaptation and cross-modal semantic enhancement to achieve superior editing quality. We present both quantitative and qualitative results, demonstrating the advantages of our approach over existing methods.

\noindent\textbf{Acknowledgments.} All experimentation and processing were conducted by the authors at the University of Rochester on the University of Rochester's servers.

%% file: main.bbl
\begin{thebibliography}{10}
\providecommand{\url}[1]{\texttt{#1}}
\providecommand{\urlprefix}{URL }
\providecommand{\doi}[1]{https://doi.org/#1}

\bibitem{dino}
Caron, M., Touvron, H., Misra, I., J{\'e}gou, H., Mairal, J., Bojanowski, P., Joulin, A.: Emerging properties in self-supervised vision transformers. In: ICCV. pp. 9650--9660 (2021)

\bibitem{chefer2023attend}
Chefer, H., Alaluf, Y., Vinker, Y., Wolf, L., Cohen-Or, D.: Attend-and-excite: Attention-based semantic guidance for text-to-image diffusion models. ACM TOG  \textbf{42}(4),  1--10 (2023)

\bibitem{chen2023disenbooth}
Chen, H., Zhang, Y., Wu, S., Wang, X., Duan, X., Zhou, Y., Zhu, W.: Disenbooth: Identity-preserving disentangled tuning for subject-driven text-to-image generation. In: The Twelfth International Conference on Learning Representations (2023)

\bibitem{vggsound}
Chen, H., Xie, W., Vedaldi, A., Zisserman, A.: Vggsound: A large-scale audio-visual dataset. In: ICASSP. pp. 721--725 (2020)

\bibitem{dhariwal2021diffusion}
Dhariwal, P., Nichol, A.: Diffusion models beat gans on image synthesis. NeurIPS  \textbf{34},  8780--8794 (2021)

\bibitem{du2023conditional}
Du, Y., Chen, Z., Salamon, J., Russell, B., Owens, A.: Conditional generation of audio from video via foley analogies. In: CVPR. pp. 2426--2436 (2023)

\bibitem{clap}
Elizalde, B., Deshmukh, S., Al~Ismail, M., Wang, H.: Clap learning audio concepts from natural language supervision. In: ICASSP. pp.~1--5 (2023)

\bibitem{avspeech}
Ephrat, A., Mosseri, I., Lang, O., Dekel, T., Wilson, K., Hassidim, A., Freeman, W.T., Rubinstein, M.: Looking to listen at the cocktail party: a speaker-independent audio-visual model for speech separation. {ACM} Trans. Graph.  \textbf{37}(4), ~112 (2018)

\bibitem{esser2021taming}
Esser, P., Rombach, R., Ommer, B.: Taming transformers for high-resolution image synthesis. In: CVPR. pp. 12873--12883 (2021)

\bibitem{text-inversion}
Gal, R., Alaluf, Y., Atzmon, Y., Patashnik, O., Bermano, A.H., Chechik, G., Cohen-Or, D.: An image is worth one word: Personalizing text-to-image generation using textual inversion. arXiv preprint arXiv:2208.01618  (2022)

\bibitem{tango}
Ghosal, D., Majumder, N., Mehrish, A., Poria, S.: Text-to-audio generation using instruction-tuned llm and latent diffusion model. arXiv preprint arXiv:2304.13731  (2023)

\bibitem{animatediff}
Guo, Y., Yang, C., Rao, A., Wang, Y., Qiao, Y., Lin, D., Dai, B.: Animatediff: Animate your personalized text-to-image diffusion models without specific tuning. arXiv preprint arXiv:2307.04725  (2023)

\bibitem{audioclip}
Guzhov, A., Raue, F., Hees, J., Dengel, A.: Audioclip: Extending clip to image, text and audio. In: ICASSP. pp. 976--980 (2022)

\bibitem{mugen}
Hayes, T., Zhang, S., Yin, X., Pang, G., Sheng, S., Yang, H., Ge, S., Hu, Q., Parikh, D.: Mugen: A playground for video-audio-text multimodal understanding and generation. In: ECCV. pp. 431--449 (2022)

\bibitem{animate-a-story}
He, Y., Xia, M., Chen, H., Cun, X., Gong, Y., Xing, J., Zhang, Y., Wang, X., Weng, C., Shan, Y., Chen, Q.: Animate-a-story: Storytelling with retrieval-augmented video generation. arXiv preprint arXiv:2307.06940  (2023)

\bibitem{prompt2prompt}
Hertz, A., Mokady, R., Tenenbaum, J., Aberman, K., Pritch, Y., Cohen{-}Or, D.: Prompt-to-prompt image editing with cross-attention control. In: ICLR (2023)

\bibitem{ddpm}
Ho, J., Jain, A., Abbeel, P.: Denoising diffusion probabilistic models. Advances in neural information processing systems  \textbf{33},  6840--6851 (2020)

\bibitem{ho2022video}
Ho, J., Salimans, T., Gritsenko, A., Chan, W., Norouzi, M., Fleet, D.J.: Video diffusion models. arXiv:2204.03458  (2022)

\bibitem{lora}
Hu, E.J., Shen, Y., Wallis, P., Allen{-}Zhu, Z., Li, Y., Wang, S., Wang, L., Chen, W.: Lora: Low-rank adaptation of large language models. In: ICLR (2022)

\bibitem{hua2023dreamtuner}
Hua, M., Liu, J., Ding, F., Liu, W., Wu, J., He, Q.: Dreamtuner: Single image is enough for subject-driven generation. arXiv preprint arXiv:2312.13691  (2023)

\bibitem{huang2023davis}
Huang, C., Liang, S., Tian, Y., Kumar, A., Xu, C.: Davis: High-quality audio-visual separation with generative diffusion models. arXiv preprint arXiv:2308.00122  (2023)

\bibitem{huan2024modeling}
Huang, C., Markovic, D., Xu, C., Richard, A.: Modeling and driving human body soundfields through acoustic primitives. arXiv preprint arXiv:2407.13083  (2024)

\bibitem{make-an-audio2}
Huang, J., Ren, Y., Huang, R., Yang, D., Ye, Z., Zhang, C., Liu, J., Yin, X., Ma, Z., Zhao, Z.: Make-an-audio 2: Temporal-enhanced text-to-audio generation. arXiv preprint arXiv:2305.18474  (2023)

\bibitem{make-an-audio}
Huang, R., Huang, J., Yang, D., Ren, Y., Liu, L., Li, M., Ye, Z., Liu, J., Yin, X., Zhao, Z.: Make-an-audio: Text-to-audio generation with prompt-enhanced diffusion models. In: ICML. pp. 13916--13932 (2023)

\bibitem{tpos}
Jeong, Y., Ryoo, W., Lee, S., Seo, D., Byeon, W., Kim, S., Kim, J.: The power of sound (tpos): Audio reactive video generation with stable diffusion. arXiv preprint arXiv:2309.04509  (2023)

\bibitem{text2video-zero}
Khachatryan, L., Movsisyan, A., Tadevosyan, V., Henschel, R., Wang, Z., Navasardyan, S., Shi, H.: Text2video-zero: Text-to-image diffusion models are zero-shot video generators. arXiv preprint arXiv:2303.13439  (2023)

\bibitem{DBLP:conf/interspeech/KilgourZRS19}
Kilgour, K., Zuluaga, M., Roblek, D., Sharifi, M.: Fr{\'{e}}chet audio distance: {A} reference-free metric for evaluating music enhancement algorithms. In: INTERSPEECH. pp. 2350--2354 (2019)

\bibitem{kong2020hifi}
Kong, J., Kim, J., Bae, J.: Hifi-gan: Generative adversarial networks for efficient and high fidelity speech synthesis. Advances in Neural Information Processing Systems  \textbf{33},  17022--17033 (2020)

\bibitem{custom-diffusion}
Kumari, N., Zhang, B., Zhang, R., Shechtman, E., Zhu, J.Y.: Multi-concept customization of text-to-image diffusion. In: CVPR. pp. 1931--1941 (2023)

\bibitem{lee2022sound}
Lee, S.H., Roh, W., Byeon, W., Yoon, S.H., Kim, C., Kim, J., Kim, S.: Sound-guided semantic image manipulation. In: CVPR. pp. 3377--3386 (2022)

\bibitem{avnerf}
Liang, S., Huang, C., Tian, Y., Kumar, A., Xu, C.: Av-nerf: Learning neural fields for real-world audio-visual scene synthesis. Advances in Neural Information Processing Systems  \textbf{36},  37472--37490 (2023)

\bibitem{liang2023neural}
Liang, S., Huang, C., Tian, Y., Kumar, A., Xu, C.: Neural acoustic context field: Rendering realistic room impulse response with neural fields. arXiv preprint arXiv:2309.15977  (2023)

\bibitem{audioldm}
Liu, H., Chen, Z., Yuan, Y., Mei, X., Liu, X., Mandic, D.P., Wang, W., Plumbley, M.D.: Audioldm: Text-to-audio generation with latent diffusion models. In: ICML. pp. 21450--21474 (2023)

\bibitem{audioldm2}
Liu, H., Tian, Q., Yuan, Y., Liu, X., Mei, X., Kong, Q., Wang, Y., Wang, W., Wang, Y., Plumbley, M.D.: Audioldm 2: Learning holistic audio generation with self-supervised pretraining. arXiv preprint arXiv:2308.05734  (2023)

\bibitem{luo2023difffoley}
Luo, S., Yan, C., Hu, C., Zhao, H.: Diff-foley: Synchronized video-to-audio synthesis with latent diffusion models (2023)

\bibitem{mo2023diffava}
Mo, S., Shi, J., Tian, Y.: Diffava: Personalized text-to-audio generation with visual alignment. arXiv preprint arXiv:2305.12903  (2023)

\bibitem{GLIDE}
Nichol, A.Q., Dhariwal, P., Ramesh, A., Shyam, P., Mishkin, P., McGrew, B., Sutskever, I., Chen, M.: {GLIDE:} towards photorealistic image generation and editing with text-guided diffusion models. In: ICML. vol.~162, pp. 16784--16804 (2022)

\bibitem{gluegen}
Qin, C., Yu, N., Xing, C., Zhang, S., Chen, Z., Ermon, S., Fu, Y., Xiong, C., Xu, R.: Gluegen: Plug and play multi-modal encoders for x-to-image generation. arXiv preprint arXiv:2303.10056  (2023)

\bibitem{clip}
Radford, A., Kim, J.W., Hallacy, C., Ramesh, A., Goh, G., Agarwal, S., Sastry, G., Askell, A., Mishkin, P., Clark, J., et~al.: Learning transferable visual models from natural language supervision. In: ICML. pp. 8748--8763 (2021)

\bibitem{t5}
Raffel, C., Shazeer, N., Roberts, A., Lee, K., Narang, S., Matena, M., Zhou, Y., Li, W., Liu, P.J.: Exploring the limits of transfer learning with a unified text-to-text transformer. The Journal of Machine Learning Research  \textbf{21}(1),  5485--5551 (2020)

\bibitem{dalle2}
Ramesh, A., Dhariwal, P., Nichol, A., Chu, C., Chen, M.: Hierarchical text-conditional image generation with clip latents. arXiv preprint arXiv:2204.06125  \textbf{1}(2), ~3 (2022)

\bibitem{stablediffusion}
Rombach, R., Blattmann, A., Lorenz, D., Esser, P., Ommer, B.: High-resolution image synthesis with latent diffusion models. In: CVPR. pp. 10684--10695 (2022)

\bibitem{unet}
Ronneberger, O., Fischer, P., Brox, T.: U-net: Convolutional networks for biomedical image segmentation. In: MICCAI. pp. 234--241 (2015)

\bibitem{mm-diffusion}
Ruan, L., Ma, Y., Yang, H., He, H., Liu, B., Fu, J., Yuan, N.J., Jin, Q., Guo, B.: Mm-diffusion: Learning multi-modal diffusion models for joint audio and video generation. In: CVPR. pp. 10219--10228 (2023)

\bibitem{dreambooth}
Ruiz, N., Li, Y., Jampani, V., Pritch, Y., Rubinstein, M., Aberman, K.: Dreambooth: Fine tuning text-to-image diffusion models for subject-driven generation. In: CVPR. pp. 22500--22510 (2023)

\bibitem{imagen}
Saharia, C., Chan, W., Saxena, S., Li, L., Whang, J., Denton, E.L., Ghasemipour, K., Gontijo~Lopes, R., Karagol~Ayan, B., Salimans, T., et~al.: Photorealistic text-to-image diffusion models with deep language understanding. NeurIPS  \textbf{35},  36479--36494 (2022)

\bibitem{instantbooth}
Shi, J., Xiong, W., Lin, Z., Jung, H.J.: Instantbooth: Personalized text-to-image generation without test-time finetuning. arXiv preprint arXiv:2304.03411  (2023)

\bibitem{make-a-video}
Singer, U., Polyak, A., Hayes, T., Yin, X., An, J., Zhang, S., Hu, Q., Yang, H., Ashual, O., Gafni, O., Parikh, D., Gupta, S., Taigman, Y.: Make-a-video: Text-to-video generation without text-video data. In: ICLR (2023)

\bibitem{song2020denoising}
Song, J., Meng, C., Ermon, S.: Denoising diffusion implicit models. In: ICLR (2020)

\bibitem{ddim}
Song, J., Meng, C., Ermon, S.: Denoising diffusion implicit models. In: ICLR (2020)

\bibitem{sound2scene}
Sung-Bin, K., Senocak, A., Ha, H., Owens, A., Oh, T.H.: Sound to visual scene generation by audio-to-visual latent alignment. In: CVPR. pp. 6430--6440 (2023)

\bibitem{any-to-any}
Tang, Z., Yang, Z., Zhu, C., Zeng, M., Bansal, M.: Any-to-any generation via composable diffusion. arXiv preprint arXiv:2305.11846  (2023)

\bibitem{van2017neural}
Van Den~Oord, A., Vinyals, O., et~al.: Neural discrete representation learning. Advances in neural information processing systems  \textbf{30} (2017)

\bibitem{wei2023elite}
Wei, Y., Zhang, Y., Ji, Z., Bai, J., Zhang, L., Zuo, W.: Elite: Encoding visual concepts into textual embeddings for customized text-to-image generation. arXiv preprint arXiv:2302.13848  (2023)

\bibitem{tune-a-video}
Wu, J.Z., Ge, Y., Wang, X., Lei, W., Gu, Y., Hsu, W., Shan, Y., Qie, X., Shou, M.Z.: Tune-a-video: One-shot tuning of image diffusion models for text-to-video generation. arXiv preprint arXiv:2212.11565  (2022)

\bibitem{xiao2023fastcomposer}
Xiao, G., Yin, T., Freeman, W.T., Durand, F., Han, S.: Fastcomposer: Tuning-free multi-subject image generation with localized attention. arXiv preprint arXiv:2305.10431  (2023)

\bibitem{diffsound}
Yang, D., Yu, J., Wang, H., Wang, W., Weng, C., Zou, Y., Yu, D.: Diffsound: Discrete diffusion model for text-to-sound generation. IEEE/ACM Transactions on Audio, Speech, and Language Processing  (2023)

\bibitem{zhao2018sound}
Zhao, H., Gan, C., Rouditchenko, A., Vondrick, C., McDermott, J., Torralba, A.: The sound of pixels. In: ECCV. pp. 570--586 (2018)

\end{thebibliography}
